\documentclass[conference]{IEEEtran}
\IEEEoverridecommandlockouts
\usepackage{cite}
\usepackage{amsmath,amssymb,amsfonts}
\usepackage{algorithmic}
\usepackage{graphicx}
\usepackage{textcomp}
\usepackage{xcolor}

\usepackage{times}  
\usepackage{helvet}  
\usepackage{courier}  
\usepackage[hyphens]{url}  
\usepackage{graphicx} 
\usepackage{caption} 

\usepackage{algorithm}
\usepackage{algorithmic}

\usepackage{subfig}
\usepackage{booktabs}
\usepackage{amsmath}
\usepackage{amssymb}
\usepackage{mathtools}
\usepackage[dvipsnames]{xcolor}
\usepackage{multirow}
\usepackage{newfloat}
\usepackage{listings}

\def\BibTeX{{\rm B\kern-.05em{\sc i\kern-.025em b}\kern-.08em
    T\kern-.1667em\lower.7ex\hbox{E}\kern-.125emX}}
\begin{document}

\title{
A Low-Rank Defense Method for Adversarial Attack on Diffusion Models
}

\author{\IEEEauthorblockN{Jiaxuan Zhu, ~~Siyu Huang*}
\IEEEauthorblockA{
\textit{Visual Computing Division} \\
\textit{School of Computing} \\
\textit{Clemson University} \\
Clemson, USA \\
\{jiaxuaz, siyuh\}@clemson.edu}
\thanks{*Corresponding author: Siyu Huang}

}

\maketitle

\begin{abstract}
Recently, adversarial attacks for diffusion models as well as their fine-tuning process have been developed rapidly. To prevent the abuse of these attack algorithms from affecting the practical application of diffusion models, it is critical to develop corresponding defensive strategies. In this work, we propose an efficient defensive strategy, named Low-Rank Defense (LoRD), to defend the adversarial attack on Latent Diffusion Models (LDMs). LoRD introduces the merging idea and a balance parameter, combined with the low-rank adaptation (LoRA) modules, to detect and defend the adversarial samples. Based on LoRD, we build up a defense pipeline that applies the learned LoRD modules to help diffusion models defend against attack algorithms. Our method ensures that the LDM fine-tuned on both adversarial and clean samples can still generate high-quality images. To demonstrate the effectiveness of our approach, we conduct extensive experiments on facial and landscape images, and our method shows significantly better defense performance compared to the baseline methods. 
\end{abstract}


\section{Introduction}

Recently, the development of diffusion models~\cite{ho2020denoising} has driven rapid progress in generative AI. Thanks to the high-quality image generation capability of the large diffusion model~\cite{rombach2022high,ramesh2022hierarchical} (e.g., latent diffusion model (LDM)~\cite{rombach2022high}), many visual generation tasks especially text-to-image generation~\cite{saharia2022photorealistic,zhang2023adding} have demonstrated excellent performance. Since the training of a large diffusion model is still not easy and very slow. To address this issue, efficient fine-tuning algorithms like DreamBooth~\cite{ruiz2023dreambooth} and LoRA~\cite{hu2021lora} for LDM and Stable Diffusion have been proposed to greatly reduce the computational efforts.

Like traditional image generation methods~\cite{goodfellow2014generative}, the rapid development of large diffusion models has also raised significant concerns about their misuse, especially in malicious editing~\cite{salman2023raising}, copyright infringement~\cite{liang2023adversarial}, and creation of fake information~\cite{van2023anti}. 
To address these concerns, 
A series of literature including Photoguard~\cite{salman2023raising}, AdvDM~\cite{liang2023adversarial}, and Anti-DreamBooth~\cite{van2023anti}  prevent target images from malicious editing. 
Although these methods can avoid the risk of abusing large diffusion models, these approaches may also be utilized for attacking diffusion models, making them be unable to perform beneficial work for people. Therefore, it is urgently necessary to design effective defense strategies to suppress the abuse of the above attacking methods. 

\begin{figure}[t]
 	\begin{center}
 		\includegraphics[width=0.6\linewidth]{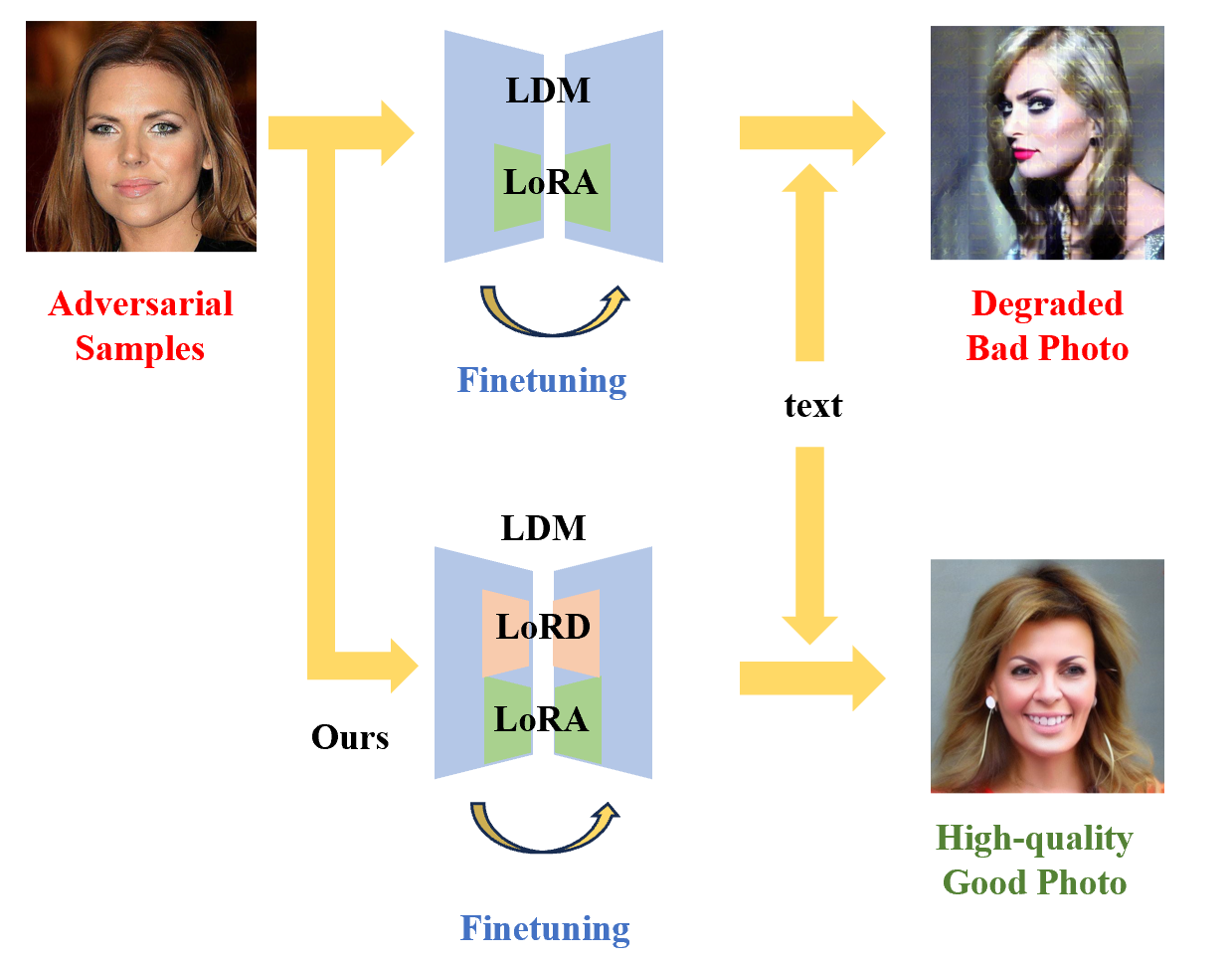}
 	\end{center}
 	\caption{Illustration of our defense strategy for Adversarial Attacks on LoRA Fine-tuning for LDM. 
  LoRD (Low-Rank Defense) is the proposed module for detecting and defending against the adversarial samples.}
 	\label{fig:idea of our defensing method}
 \end{figure}

Among the attacking methods, ACE/ACE+ attacking in mist-v2~\cite{zheng2023understanding} is important, as it can interfere with the LoRA fine-tuning process of the large diffusion models, such that it might also hinder the training process of diffusion models. In addition, defending against this type of attack algorithm is very challenging, as traditional adversarial defense methods mainly focus on the attacks on model inputs rather than the training process of the target models. Therefore, in this work we aim to design a defense strategy to defend against the ACE/ACE+ attacks. Our crucial idea could be illustrated as Fig.~\ref{fig:idea of our defensing method}.
Specifically, we propose the Low-Rank Defense (LoRD) module (see Fig.~\ref{fig:Structure of our defensing LoRA}),
which introduces a LoRA module to defend against the adversarial samples, and additionally utilizes a multilayer perceptron (MLP) to obtain a balance parameter for merging the original LoRA branch into LoRD, such that our LoRD can not only detect and defend against the adversarial samples, but also fine-tune on the clean samples.

Based on this essential LoRD module, we build a defense pipeline, as shown in Fig.~\ref{fig:pipeline}, to defend the ACE/ACE+ attacks in real-world scenarios. This pipeline includes two stages: Stage-1 learns the LoRD modules by refering to the idea of adversarial training~\cite{shafahi2019adversarial}; Stage-2 firstly merges the LoRD weights into the LDM and then optimizes the original LoRA using adversarial samples and/or clean samples. In the testing phase, our pipeline merges both LoRD and original LoRA into the LDM to ensure high-quality text-to-image generations with either attacked or clean image samples.

We conduct extensive experiments on facial and landscape image datasets to validate our approach. Qualitative and quantitative results demonstrate that our defense method greatly alleviates the impact of adversarial attacks on the target model, and it is significantly more effective in defending against this type of attacks than traditional adversarial training ideas.
We summarize the contributions of this work as follow:
\begin{enumerate}
    \item This work provides a novel idea for studying the problem of defensing adversarial attacks against LoRA fine-tuning of LDMs. 
    \item We propose a Low-Rank Defense (LoRD) module to efficiently defend the ACE/ACE+ attack LoRD actually utilizes LoRA. LoRD introduces the merging idea of LoRA modules and a balance parameter to ensure it can be applied to both clean and adversarial samples.
    \item We set up an effective defense pipeline for diffusion model fine-tuning. Extensive text-to-image generation experiments on facial and landscape datasets demonstrate the effectiveness of our pipeline.
\end{enumerate}

\section{Related Work}
\subsection{Diffusion Models and Their Fine-tuning Strategies}
Recently, diffusion-based models~\cite{ho2020denoising} present outstanding performances in image generation tasks.
Zhang et al.~\cite{zhang2023add} offers ControlNet to realize the conditional text-to-image generations. Rombach et al.~\cite{rombach2022high} proposes Latent Diffusion Model (LDM) to acquire high-resolution generation images. Our work mainly solves the robustness issue of LDM. Thanks to the LAION-5B dataset~\cite{schuhmann2022laion}, Rombach et al.~\cite{rombach2022high} upgrade its LDM to Stable Diffusion~\footnote{~\url{https://github.com/CompVis/stable-diffusion}}, which offers higher resolution generation images and introduce CLIP models~\cite{radford2021learning} as text encoders. 
Due to the large Stable Diffusion model, some fine-tuning methods ~\cite{hu2021lora,ruiz2023dreambooth} for Stable Diffusion have been proposed. Among them, DreamBooth~\cite{ruiz2023dreambooth} provides a feasible method for customizing outputs. LoRA for Stable Diffusion~\footnote{~\url{https://github.com/cloneofsimo/lora}} provides a solution that can change the model output by only fine-tuning a few parameters of Stable Diffusion. 
Recently, adversarial attack methods on LDM's LoRA fine-tuning has been proposed~\cite{zheng2023understanding}, it might do severe harm to the normal customization process of LDM. This work aims to address this challenge by developing efficient defensive strategies.

\subsection{Adversarial Attack}
The study on adversarial robustness originates from Fast Gradient Sign Method (FGSM)~\cite{goodfellow2014explaining}, which applies adversarial perturbations to implement adversarial attacks on deep neural networks. Inspired by FGSM, many white-box attack methods have been proposed~\cite{madry2017towards,moosavi2017universal,su2019one,xiao2018generating,moosavi2016deepfool}. Among them, Projected Gradient Descent (PGD)~\cite{madry2017towards} is one of the very efficient attack methods that it adopts the iterative attacking idea. In addition to these white-box attacks, the black-box attacks~\cite{papernot2017practical,liu2016delving} have also been explored, which would require access to little or even no information of the deep learning models.

For adversarial robustness of Diffusion Models, Anti-DreamBooth~\cite{van2023anti} successfully attacks DreamBooth fine-tuning for Stable Diffusion, AdvDM in mist-v1~\cite{liang2023adversarial} and Photoguard~\cite{salman2023raising} proposed adversarial attacks for diffusion models. More recently, ACE/ACE+ attacking of mist-v2~\cite{zheng2023understanding} provides an efficient algorithm for interfering with LoRA fine-tuning for LDM. 
Our work aims to present a pipeline to defend adversarial attacks on LoRA fine-tuning of LDMs.

\subsection{Adversarial Defense} Adversarial defense~\cite{liao2018defense} aims to prevent deep learning models from being attacked by the adversarial attack methods. As one of the promising directions, adversarial training~\cite{shafahi2019adversarial} generates adversarial samples using the attack methods and train the target models using adversarial samples and clean samples together, thereby achieving the goal of adversarial defense. Our work also falls under the scope of adversarial training by adopting the idea of low-rank adaptation to efficiently fine-tune the diffusion models on adversarial samples.

\section{Methodology}
\subsection{Problem Setting}
In this work, we study the problem of protecting Latent Diffusion Model (LDM)'s LoRA~\cite{hu2021lora} fine-tuning from being destroyed by attacking methods such as mist-v2~\cite{zheng2023understanding}. More specifically, for a pre-trained LDM~\cite{rombach2022high} $f\left( \cdot \right)$ which consists of a pair of encoder $\mathcal{E} \left( \cdot \right) $ and decoder $\mathcal{D} \left( \cdot \right)$ with parameters $\theta$. 
Given an image-text paired dataset $\left\{ \mathcal{X} ,\mathcal{Y} \right\}$, $\mathcal{X}$ denotes a collection of images in this dataset and $\mathcal{Y}$ denotes a collection of prompts. For each image $x\in \mathcal{X}$, an adversarial attack adds a small perturbation $\delta$ onto the original image $x$, obtaining the adversarial sample $x^{\prime}=x+\delta$ to deceive the LoRA fine-tuning process for the target LDM $f\left( \cdot \right)$. 
For instance, mist-v2 attacks LDM with Consistent Errors (ACE)~\cite{zheng2023understanding} for obtaining $x^{\prime}$. To defend diffusion model attack, the defense method should ensure the defensed target model $f^{\prime}\left( \cdot \right)$ can be fine-tuned with adversarial samples while outputing high-quality outputs, as shown as Fig.~\ref{fig:idea of our defensing method}.

\subsection{Preliminary: Low-Rank Adaptation (LoRA)}
For the standard LoRA, given a pre-trained weight matrix $W\in \mathbb{R}^{d\times d}$, the output of the LoRA is:
\begin{small}
\begin{equation}
    \begin{aligned}
    \label{eq:Standard LoRA}
        W^{\prime}=W+\bigtriangleup W=W+BA
    \end{aligned}
\end{equation}
\end{small}
$\bigtriangleup W = BA$ denotes low-rank matrices of the LoRA, where $B\in \mathbb{R} ^{d\times r},A\in \mathbb{R} ^{r\times d}$, and the rank $r\ll d$~\cite{hu2021lora}. The ouptut of the LoRA and LDM is:
\begin{small}
\begin{equation}
    \begin{aligned}
    \label{eq:Standard LoRA ouput}
        h_0=Wx+\bigtriangleup Wx=Wx+BAx
    \end{aligned}
\end{equation}
\end{small}
In the fine-tuning process, $\bigtriangleup Wx$ would be scaled by parameters $\frac{\alpha}{r}$. Thus, final output of original LoRA and LDM is:
\begin{small}
\begin{equation}
    \begin{aligned}
    \label{eq:Standard Final LoRA ouput}
        h_0=Wx+\frac{\alpha}{r}\bigtriangleup Wx=Wx+\frac{\alpha}{r}BAx, \, \alpha \geqslant r
    \end{aligned}
\end{equation}
\end{small}

\subsection{Low-Rank Defense (LoRD)}
To address the challenging problem of defensing adversarial attack on LoRA-based LDM fine-tuning, we introduce an efficient defensing module named Low-Rank Defense (LoRD), preventing diffusion model fine-tuning process from being affected by adversarial examples. The general idea of LoRD is adopting a two-branch LoRA design to make diffusion model work for both clean and adversarial sample, as well as learning a balance parameter to integrate these two branches. Thanks to its low-rank manner, LoRD is effective for adversarial defense yet computationally efficient.

\begin{figure}[t]
 	\begin{center}
 		\includegraphics[width=0.7\linewidth]{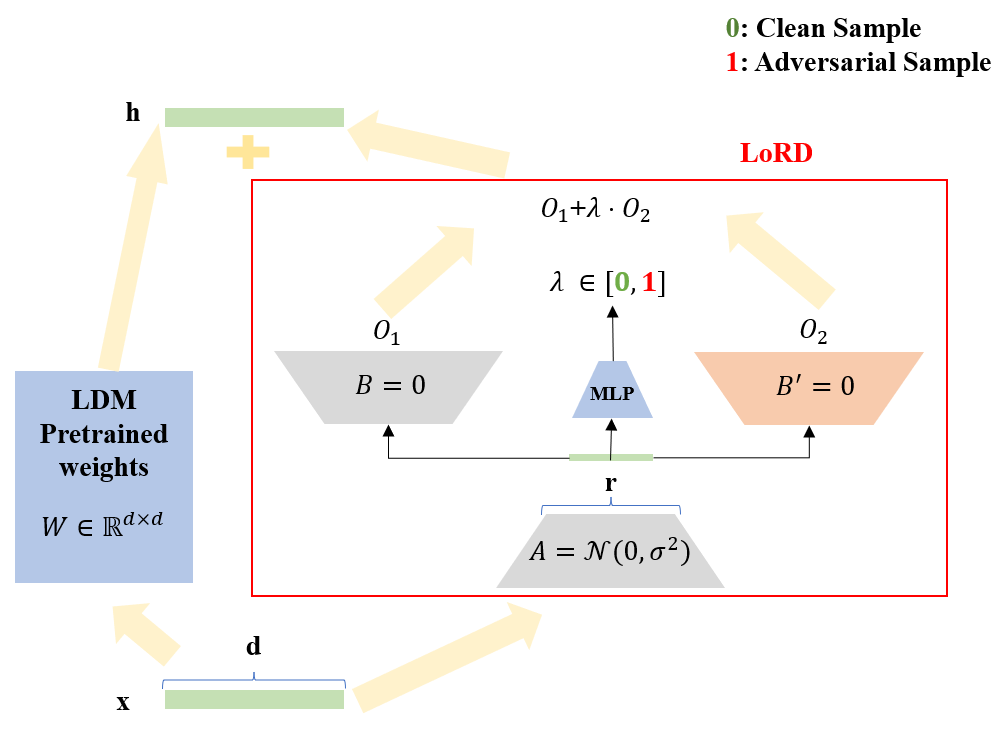}
 	\end{center}
 	\caption{An overview of our Low-Rank Defense (LoRD). Gray part denotes the original LoRA. Orange part denotes the second LoRA branch specifically learned for adversarial defense. $\lambda$ is optimized by the BCE Loss in Eq.~\ref{eq:adversarial training}.}
 	\label{fig:Structure of our defensing LoRA}
 \end{figure}

Specifically, as illustrated in Fig.~\ref{fig:Structure of our defensing LoRA}, compared to the standard LoRA, our LoRD method introduces one more MLP module and one more matrix $B^{\prime}$. 
The MLP module consists of an MLP head $MLP\left( \cdot \right)$ and a Sigmoid function $\sigma(\cdot)$ to calculate the balancing parameter $\lambda$, as
\begin{small}
\begin{equation}
    \begin{aligned}
    \label{eq:Balance_r}
        \lambda&=\sigma(MLP(\frac{\alpha}{r}\bigtriangleup Wx))\\&=\sigma(MLP(\frac{\alpha}{r}BAx))
    \end{aligned}
\end{equation}
\end{small}
Balance parameter $\lambda$ estimates the probability of an input feature $x$ being from a clean or an adversarial sample, balancing between matrices $B$ and $B^{\prime}$ for LoRD output. 

The output of standard LoRA is $O_1=\frac{\alpha}{r}BAx$. Our LoRD method introduces another matrix $B^{\prime}$ to obtain $O_2=\frac{\alpha}{r}B^{\prime}Ax$. $B^{\prime}$ denotes the second LoRA branch in LoRD module, which is learned to handle attacked samples.

Finally, its output is a weighted merge of $O_1$ and $O_2$, as
\begin{small}
\begin{equation}
    \begin{aligned}
    \label{eq:Our Final LoRA ouput}
        h&=Wx+\bigtriangleup W^{\prime}x\\&=Wx+\left( O_1+\lambda\cdot O_2 \right) x\\&=Wx+\frac{\alpha}{r}BAx+\lambda\cdot \frac{\alpha}{r}B^{\prime}Ax
    \end{aligned}
\end{equation}
\end{small}
where $h$ denotes the final output of LoRD, $\lambda$ is the balance parameter for merging two branch outputs $O_1$ and $O_2$. In addition to the original branch $O_1$, $O_2$ provides the low-rank structure to defend the adversarial attacks. Thanks to the balance parameter $\lambda$, our LoRD can decide to what extent the defense function of the second branch $O_2$ should be embedded into the final output $h$ according to the input sample. Ideally, if the input sample $x$ is a clean sample, the value of the balance parameter $\lambda$ should be close to $0$, so the output LoRD is closer to that of the original LoRA, and the defense function of the second branch $O_2$ is introduced as little as possible; if the input sample $x$ is an adversarial sample, LoRD should defend the attacked sample as well as possible, so the value of the balance parameter $\lambda$ is close to $1$, and the defense function of the second branch $O_2$ will be utilized by LoRD maximumly. These designs all promote our LoRD to defend the attacks for LoRA fine-tuning efficiently.

\subsection{An Adversarial Defense Pipeline for Diffusion Models}

  \begin{figure*}[t]
 	\begin{center}
 		\includegraphics[width=0.8\linewidth]{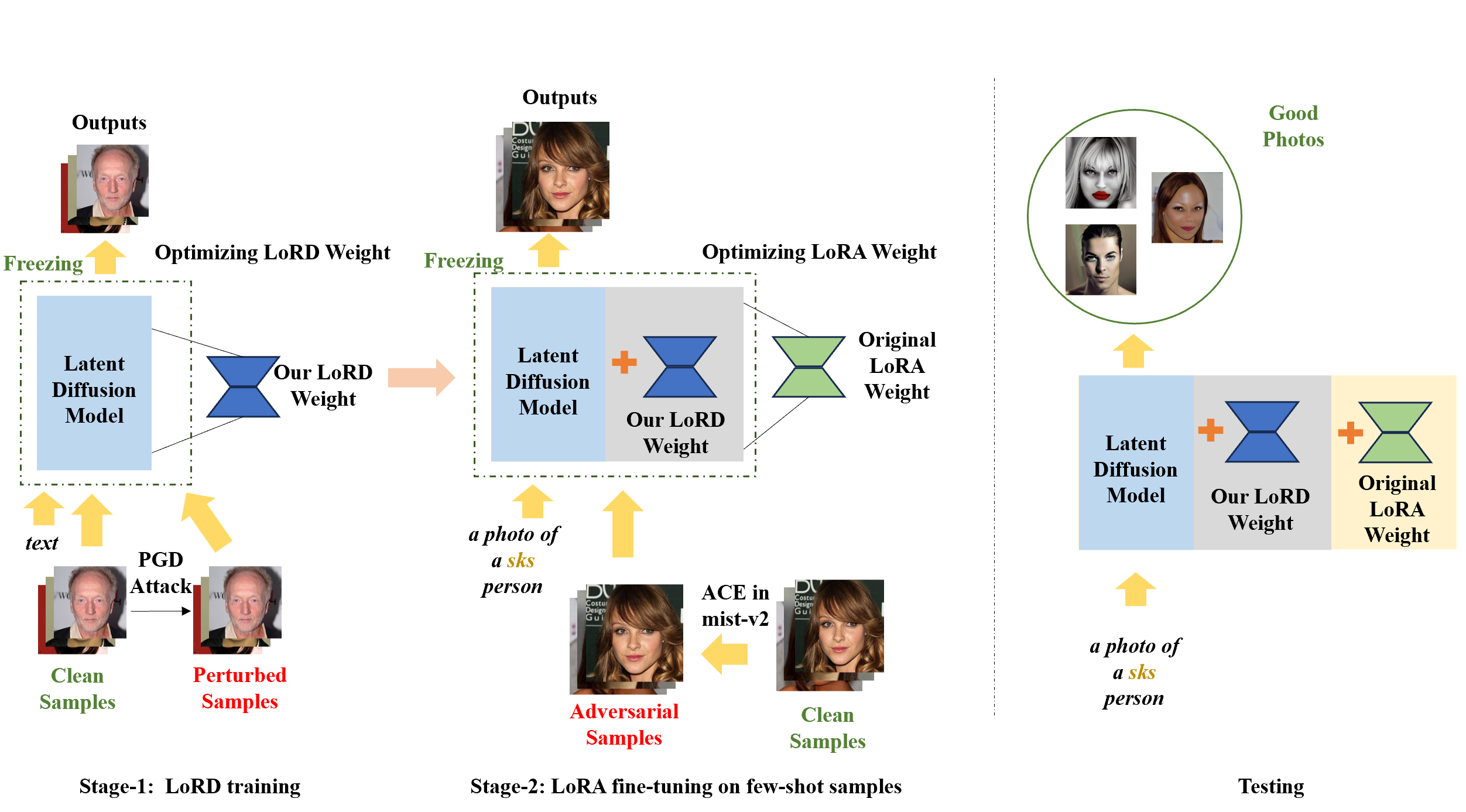}
 	\end{center}
 	\caption{Overview of our defense pipeline. In Stage-1, we utilize clean samples in training dataset to generate perturbed samples and corresponding texts, and fine-tune LoRD weight with Eq.~\ref{eq:adversarial training}. In Stage-2, we introduce the pretrained weight of LoRD and then fine-tune the LoRA module with adversarial samples according to Eq.~\ref{eq:LDM loss stage-2}. In the testing phase, we merge the LoRD weight and LoRA weight together with the pretrained LDM to generate high-quality output images according to the prompts.
  }
 	\label{fig:pipeline}
 \end{figure*}
 
Based on the proposed LoRD module, we build up the defense pipeline for diffusion model finetuning process, as illustrated in Fig.~\ref{fig:pipeline}. This pipeline contains two stages in training, as follows.

\subsubsection{Stage-1: LoRD training.}
In Stage-1, we employ image-text data to learn the LoRD modules. Given an image-text dataset $\left\{ \mathcal{X} ,\mathcal{Y} \right\}$, we  utilize the PGD attack~\cite{madry2017towards} and refer the traditional adversarial attack idea from photoguard~\cite{salman2023raising} and AdvDM~\cite{liang2023adversarial} to generate the corresponding perturbed images $x_{per}$ for each image $x \in \mathcal{X}$ and its text $y \in \mathcal{Y}$, as
\begin{small}
\begin{equation}
    \begin{aligned}
    \label{eq:PGD attack}
        \begin{cases}
        	x^{i+1}=\prod_{x+S}{\left( x^i+\alpha _{per}sign\left( \bigtriangledown _x\mathcal{L} _{LDM} \right) \right)}\\
        	x_{per}=x^C, 0\leqslant i<C\left( C=1,2,3,... \right)\\
            \mathcal{L} _{LDM} \coloneqq \mathbb{E} _{\varepsilon \left( x \right) ,\epsilon \thicksim \mathcal{N} \left( 0,1 \right) ,t}\left[ \left\| \left. \epsilon -\epsilon _{\theta}\left( z_t,t \right) \right\| _{2}^{2} \right. \right] \\
            \left\| \left. x_{per}-x \right\| \right. \leqslant \zeta \\
        \end{cases}
    \end{aligned}
\end{equation}
\end{small}
where $C$ represents the number of iterations for attacks, $S$ means the boundary for perturbations, $\alpha _{per}$ is the step length of the attack, $\zeta$ is the size limitation of perturbations. $\mathcal{L} _{LDM}$ denotes the loss function of LDM, $\varepsilon \left( x \right)$ represents our given LDM encoder, neural backbone $\epsilon _{\theta}\left( z_t,t \right)$ is a time-conditional UNet and $z_t$ is obtained by the given encoder $\varepsilon \left( x \right)$ during the training process~\cite{rombach2022high}. After obtaining perturbed image $x_{per}$, we learn LoRD modules by adversarial training as
\begin{small}
\begin{equation}
    \begin{aligned}
    \label{eq:adversarial training}
        \mathcal{L} _{Def}=\mathcal{L} _{LDM} + \lambda_{adv}*\mathcal{L}^{adv} _{LDM} + \lambda_{det}*BCELoss\left( \lambda, labels \right)
    \end{aligned}
\end{equation}
\end{small}
With this objective function, we feed both clean samples and perturbed samples into the model together for training. 
$\lambda_{adv}$ weights the adversarial training term, $\mathcal{L}_{LDM}$ is optimized on the clean sample $x$ as illustrated in Eq. \ref{eq:PGD attack}, $\mathcal{L}^{adv}_{LDM}$ is optimized on the perturbed sample $x_{per}$ as:
\begin{small}
\begin{equation}
    \begin{aligned}
    \label{eq:adversarial LDM loss}
        \mathcal{L}^{adv} _{LDM} \coloneqq \mathbb{E} _{\varepsilon \left( x_{per} \right) ,\epsilon \thicksim \mathcal{N} \left( 0,1 \right) ,t}\left[ \left\| \left. \epsilon -\epsilon _{\theta}\left( z^{per}_t,t \right) \right\| _{2}^{2} \right. \right]
    \end{aligned}
\end{equation}
\end{small}
where $z^{per}_t$ is the corresponding latent for $x_{per}$. 
Balance parameter $\lambda$ is optimized using BCE Loss, $labels$ is the corresponding labels (zero for the clean sample, one for the perturbed sample), and $\lambda_{det}$ denotes the weighting parameter for this term. 

\subsubsection{Stage-2: LoRA fine-tuning.}
Stage-2 learns the standard LoRA module based on the merge of pre-trained LDM weights and the LoRD weights learned in Stage-1. In this stage, the input prompt is added a key string \textit{sks} to meet the requirements of personalization. Stage-2 is fine-tuned with few-shot adversarial samples $x^{\prime}$, as:
\begin{small}
\begin{equation}
    \begin{aligned}
    \label{eq:LDM loss stage-2}
        \mathcal{L}^{'} _{LDM} \coloneqq \mathbb{E} _{\varepsilon \left( x^{'} \right) ,\epsilon \thicksim \mathcal{N} \left( 0,1 \right) ,t}\left[ \left\| \left. \epsilon -\epsilon _{\theta}\left( z^{'}_t,t \right) \right\| _{2}^{2} \right. \right]
    \end{aligned}
\end{equation}
\end{small}
where $z^{'}_t$ is corresponding latent for adversarial sample $x^{'}$. 

After Stage-2 training, we merge the pre-trained LDM, LoRD, and LoRA as:
\begin{small}
\begin{equation}
    \begin{aligned}
    \label{eq:merging original LoRA}
        h^{\prime}&=\left( W+\bigtriangleup W^{\prime} \right) x^{\prime}+\bigtriangleup W^{''}x^{\prime}\\&=\left( W+\left( O_1+\lambda\cdot O_2 \right) \right) x^{\prime}+\frac{\alpha ^{''}}{r^{''}}B^{''}A^{''}x^{\prime}\\&=Wx^{\prime}+\frac{\alpha}{r}BAx^{\prime}+\lambda\cdot \frac{\alpha}{r}B^{\prime}Ax^{\prime} + \frac{\alpha ^{''}}{r^{''}}B^{''}A^{''}x^{\prime}
    \end{aligned}
\end{equation}
\end{small}
where $h^{'}$ is the final output, $\bigtriangleup W^{''}$ is the original LoRA weight which consists of matrixes $B^{''},A^{''}$, and $\alpha ^{''} \geqslant r^{''}$.

\subsubsection{Testing phase.}
The testing phase of our robust LDM is illustrated in the right part of Fig.~\ref{fig:pipeline}. The key string \textit{sks} is embedded into the prompt for LoRA personalisation. Since LoRD is fine-tuned on both clean and adversarial samples, the LDM model is able to generate high-quality images for either clean or adversarial inputs.

\begin{figure}[!t]
\begin{center}
    \subfloat[CelebA-HQ]{
        \begin{minipage}{\linewidth}
        \centering
        \includegraphics[width=0.9\linewidth]{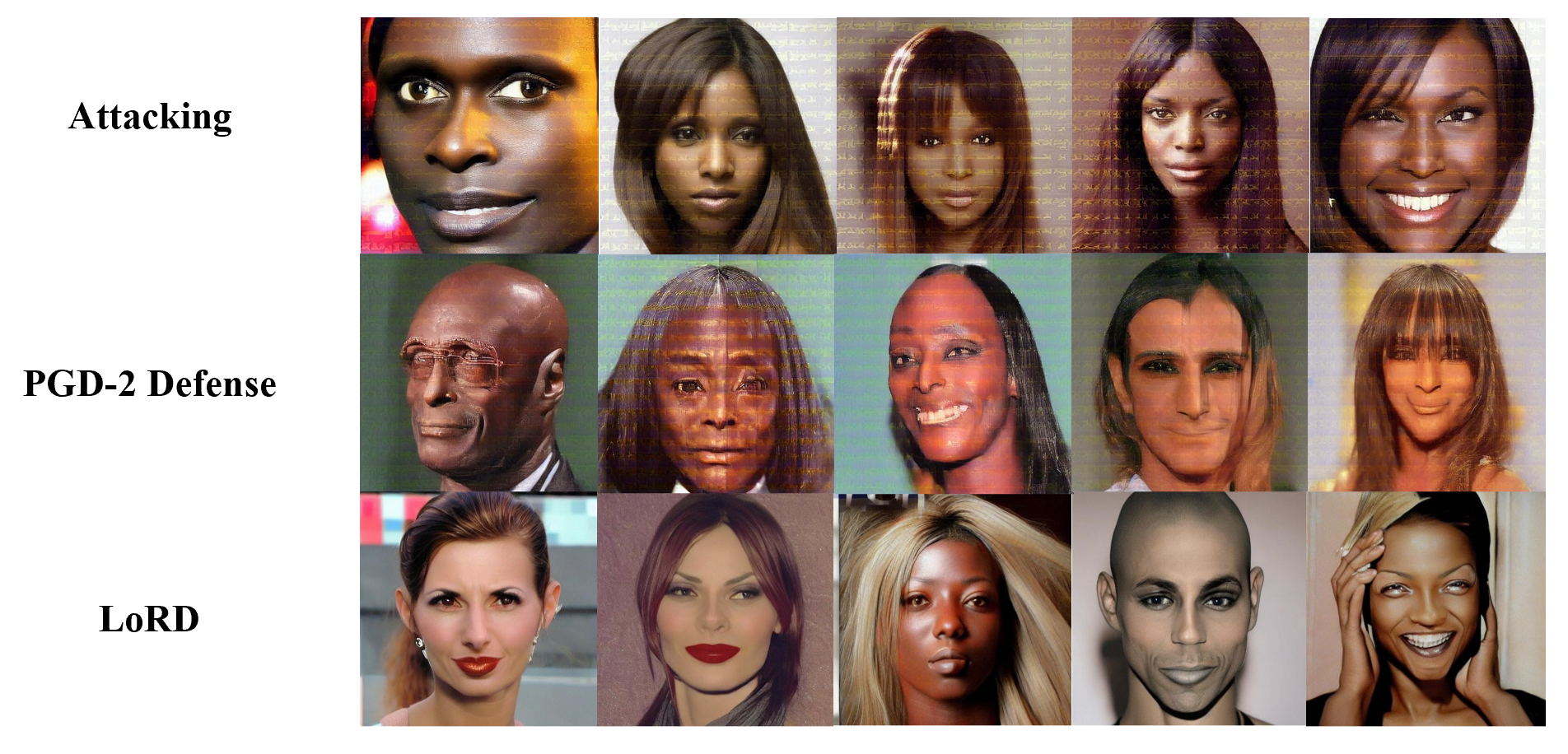}
        \label{fig:Defense for mist-v2 adversarial samples for Face:a}
        \end{minipage}
    }
    \\
    \subfloat[VGGFace2]{
        \begin{minipage}{\linewidth}
        \centering
        \includegraphics[width=0.9\linewidth]{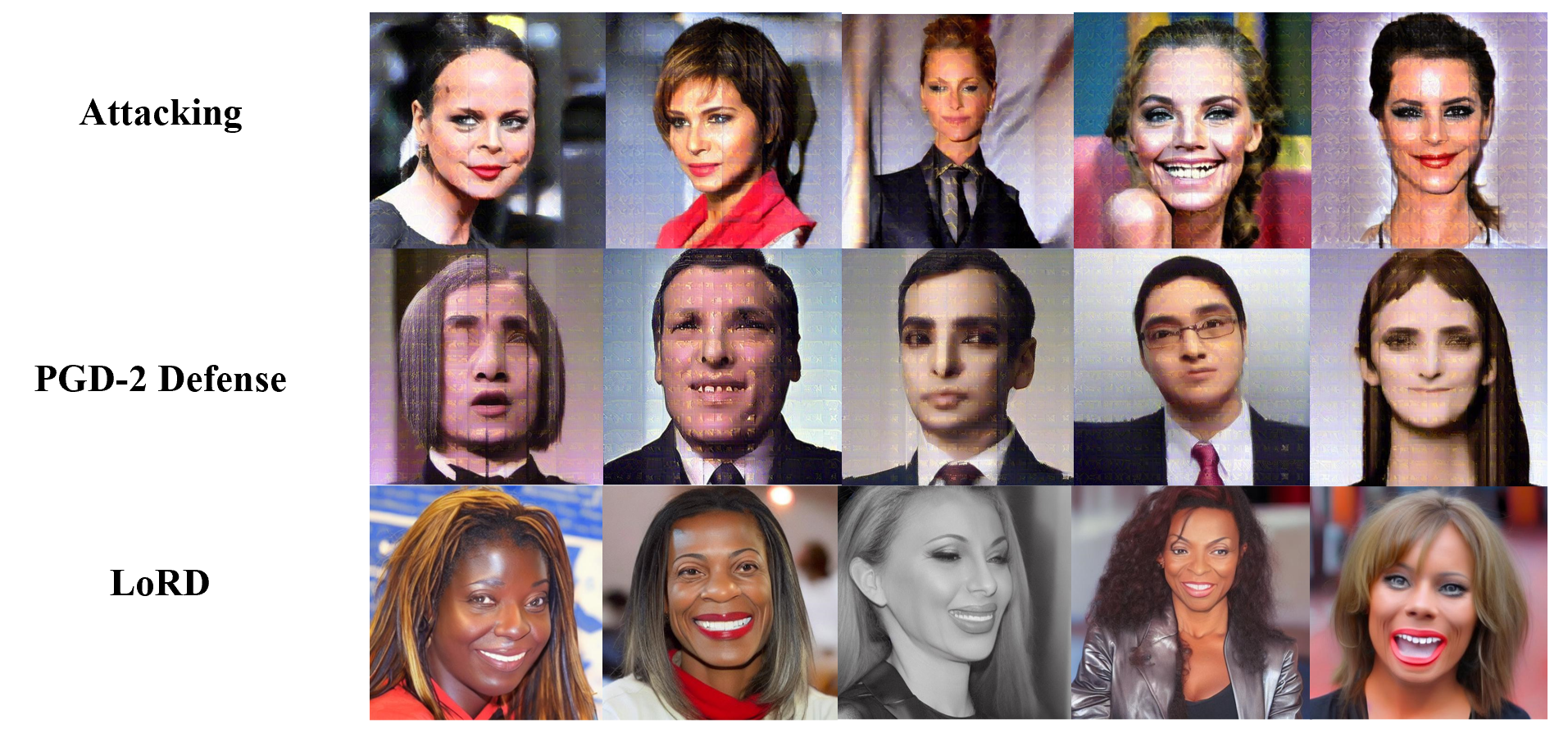}
        \label{fig:Defense for mist-v2 adversarial samples for Face:b}
        \end{minipage}
    }
\end{center}
   \caption{Comparisons among our defense method, adversarial training defense method using PGD-2 and LoRA fine-tuning, and ACE attacking in mist-v2 on face images.}
\label{fig:Defense for mist-v2 adversarial samples for Face}
\end{figure}

\begin{table}[!t]
  \centering
  \caption{Quantitative results on the face and landscape data.}
  \vspace{-.5em}
  \resizebox{1.0\columnwidth}{!}{%
    \begin{tabular}{c|cccccc}
    \toprule
    \multirow{2}[4]{*}{} & \multicolumn{2}{c}{CelebA-HQ} & \multicolumn{2}{c}{VGGFace2} & \multicolumn{2}{c}{Landscape} \\
\cmidrule{2-7}          & CLIP-IQA $\uparrow$ & FID $\downarrow$   & CLIP-IQA $\uparrow$ & FID $\downarrow$   & CLIP-IQA $\uparrow$ & FID $\downarrow$ \\
    \midrule
    Attacking & 0.362  & 154.161  & 0.385  & 182.771  & 0.607  & 133.289  \\
    PGD-2 defense & 0.485  & 146.686  & 0.203  & 152.525  & 0.240  & 150.136  \\
    LoRD  & \textbf{0.493 } & \textbf{100.727 } & \textbf{0.620 } & \textbf{137.918 } & \textbf{0.675 } & \textbf{95.262 } \\
    \bottomrule
    \end{tabular}%
    }
  \label{fig:Main Quantitative Results}%
\end{table}%

 \begin{figure}[t]
 	\begin{center}
 		\includegraphics[width=0.8\linewidth]{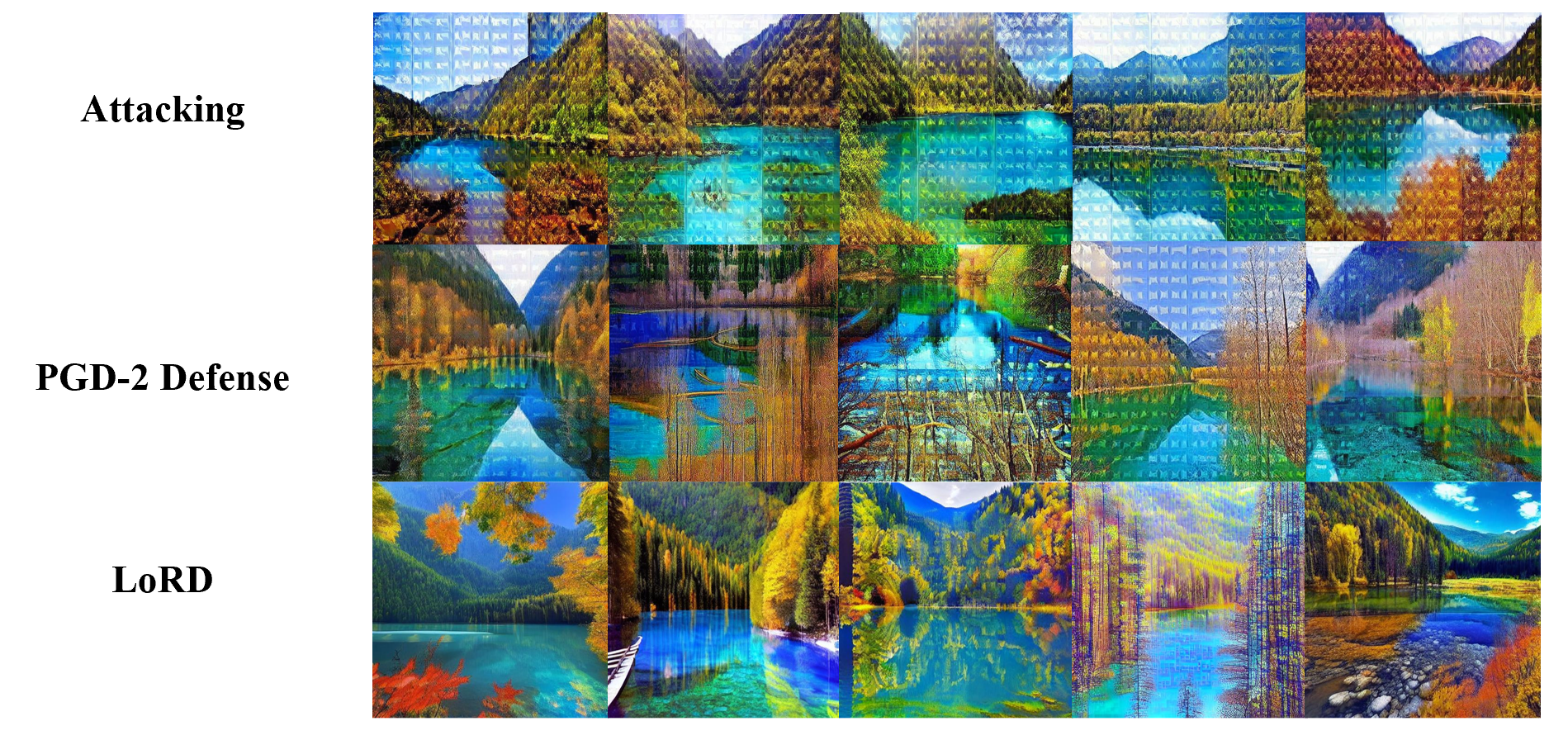}
 	\end{center}
 	\caption{Comparisons among our defense method, adversarial training defense method using PGD-2 and LoRA fine-tuning, and ACE attacking in mist-v2 on landscape images.}
 	\label{fig:Defense for mist-v2 adversarial samples for Landscape}
 \end{figure}

\begin{figure}[!t]
\begin{center}
    \subfloat{
        \begin{minipage}{\linewidth}
        \centering
        \includegraphics[width=0.8\linewidth]{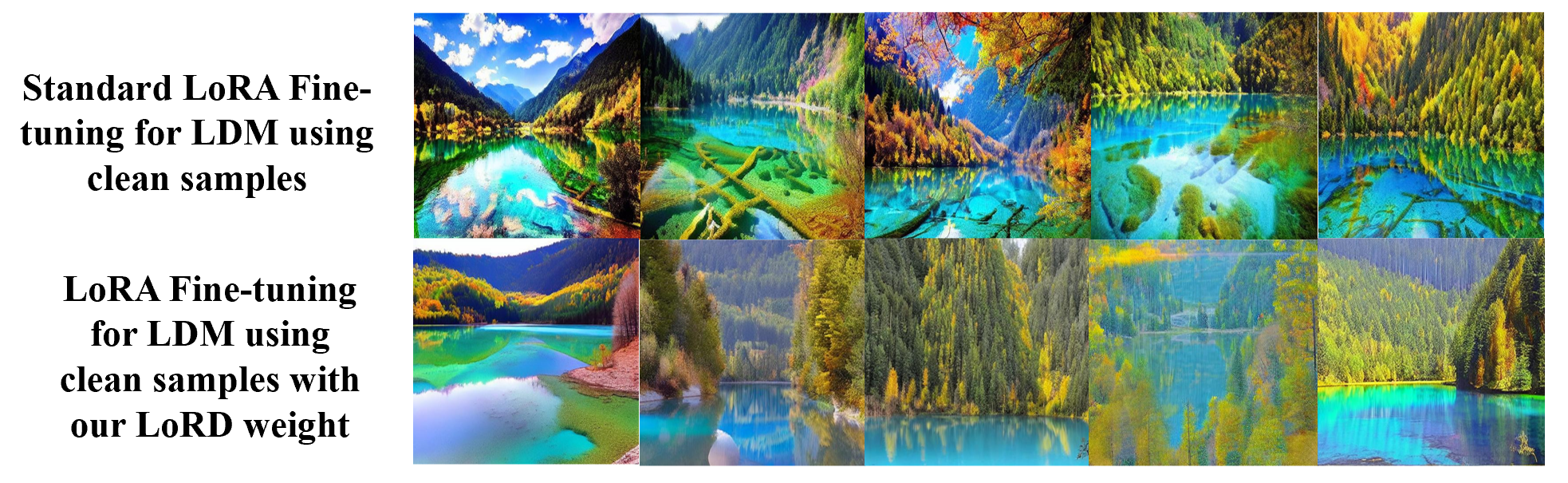}
        \label{fig:Clean samples finetuning for Face:b}
        \end{minipage}
    }
\end{center}
   \caption{Visualizations of LoRA fine-tuning results for LDM using clean images.}
\label{fig:Clean samples finetuning for Face}
\end{figure}

\section{Experiments}

\subsection{Experimental Setups}
\subsubsection{Datasets.}
Our training datasets for LoRD training in Stage-1 include CelebA-HQ-caption~\footnote{~\url{https://huggingface.co/datasets/Ryan-sjtu/celebahq-caption}}, 
and Chinese$\_$Landscape$\_$Paintings$\_$1k~\footnote{~\url{https://huggingface.co/datasets/mingyy/chinese_landscape_paintings_1k}} collected from Hugging Face. The former dataset evaluate the defense performance on face images, and the latter dataset evaluates on landscape images. For the LoRA fine-tuning in Stage-2, 100 test images are randomly sampled from CelebA-HQ and VGGFace2 dataset in Anti-DreamBooth project~\cite{van2023anti}~\footnote{~\url{https://github.com/VinAIResearch/Anti-DreamBooth}} for evaluation on the face task, and we collect another 100 test images from websites~\footnote{~\url{https://www.bing.com/images/search?q=jiuzhaigou&qs=n&form=QBIR&sp=-1&ghc=1&lq=0&pq=jiuzhaigou&sc=10-10&cvid=313F4EAF5B7942108A381382F6F1963C&ghsh=0&ghacc=0&first=1}} for evaluation on the landscape task.

\subsubsection{Implementation details.}
Our expereimental target is LDM and utilize Stable Diffusion model to validate our defense method. We adopt the state-of-the-art attacking baseline, ACE attacking~\cite{zheng2023understanding} from mist-V2, to validate the defense results of our method. The target image downloaded from mist-v2 projects~\footnote{~\url{https://github.com/psyker-team/mist-v2/blob/main/data/MIST.png}}. We follow mist-v2 setting~\cite{zheng2023understanding} to set the iterations as $N=4, M=10, K=50$, step length as $\alpha^{\prime}=5\times 10^{-3} $, learning rate as $1\times 10^{-5}$, $\zeta$ as $8/255$, $4/255$ and $16/255$ for CelebA-HQ, VGGface2 and landscape images. 

For our LoRD and stage-1 in our pipeline, we select 1,000 images in each training dataset and utilize the PGD attacks with the same parameters as Anti-DreamBooth~\cite{van2023anti}, using only $2$ iterations to generate perturbed images. For the objective function, we set $\lambda_{adv}=2$ and $\lambda_{det}=0.1$ and calculate $BCELoss$ with clean samples and perturbed samples separately and add the two parts to get the average losses for $\lambda$. The training epoch is $100$, learning rate is $1\times10^{-4}$, and the parameters $\alpha, r$ for LoRD are $32$ and $4$, respectively. Other seetings are the same as the LoRA in diffusers~\footnote{~\url{https://github.com/huggingface/diffusers/tree/main/examples/research_projects/lora}}. The LoRD weight trained on CelebA-HQ-caption dataset is used for validating face images including 100 test images from CelebA-HQ dataset and 100 test images from VGG-Face2 dataset. The LoRD weight trained on chinese$\_$landscape$\_$paintings$\_$1k is used for validating the collected 100 landscape images.

For the original LoRA and stage-2 in our pipeline, we also follow the mist-v2 setting~\cite{zheng2023understanding} to combine LoRA and DreamBooth~\cite{ruiz2023dreambooth} to obtain the personalized LDM. The personalized prompts are designed as \textit{a photo of a sks person} for the face task and \textit{a photo of a sks landscape} for the landscape task.

For the testing stage, we load the trained LoRD weight and LoRA weight. We input the prompts of \textit{a photo of a sks person} and \textit{a photo of a sks landscape} to separately generate 100 images to validate our defense method.



\subsubsection{Evaluation metrics.}
We employ FID~\cite{heusel2017gans} and CLIP-IQA~\cite{wang2022exploring} to assess the image quality of the output images. For CLIP-IQA, we utilize the prompts of \textit{A good photo of a person} and \textit{A bad photo of a person} for the face task, another group prompts \textit{A good photo of a landscape} and \textit{A bad photo of a landscape} for the landscape task. Tthe higher the CLIP-IQA score, the closer it is to a good quality image. We calculate FID from generated images in testing phase and clean samples in Stage-2, and lower score means better image generations. In addition, we introduce the PGD-2 attacks as the comparing object, we would apply original LoRA to fine-tune the Stable Diffusion model using both clean samples and adverial samples generated from PGD-2 attacks. This comparing method is essentially fine-tuning the target model using only one branch of our LoRD. For the consideration of convenience, we still name this method as \textit{PGD-2 defense} and will slightly abuse this name in the following text.

 \begin{figure}[!t]
        \subfloat[CLIP-IQA]{
        \begin{minipage}{0.5\linewidth}
        \centering
        \includegraphics[width=\linewidth]{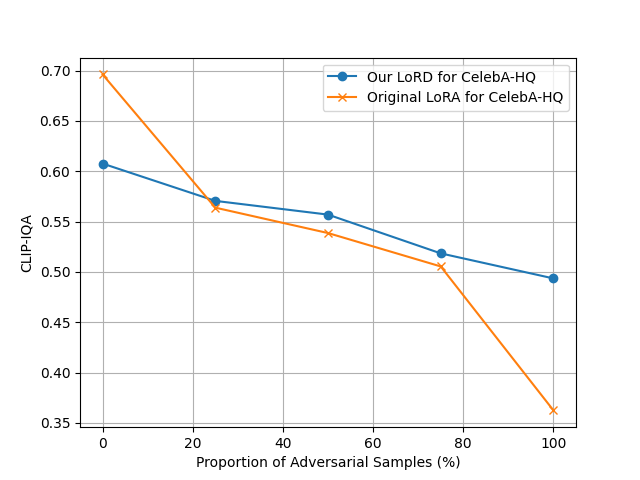}
        \label{fig:mixed_results_merged:a}
        \end{minipage}
        }
        \subfloat[FID]{
            \begin{minipage}{0.5\linewidth}
            \centering
            \includegraphics[width=\linewidth]{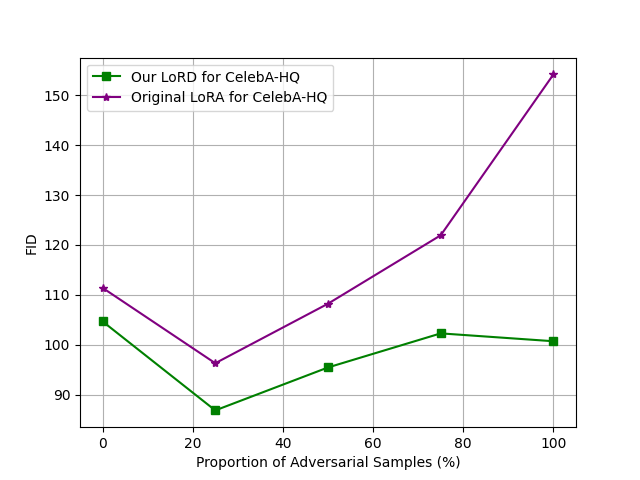}
            \label{fig:mixed_results_merged:b}
            \end{minipage}
        }
 	\caption{Comparisons of quantitative results between our LoRD pipeline and original LoRA for fine-tinued mixed CelebA-HQ testing images.} 
 	\label{fig:mixed_results_merged}
 \end{figure}

\subsection{Qualitative Results}

We visualize attacking results and our defensing results in Fig.~\ref{fig:Defense for mist-v2 adversarial samples for Face} and Fig.~\ref{fig:Defense for mist-v2 adversarial samples for Landscape}. These figures demonstrate the ACE attack in mist-v2 can greatly affect the LoRA fine-tuning process of LDM, and adversarial fine-tuning using PGD-2 may not defend this attack well. Comparing to adversarial training using PGD-2 and LoRA fine-tuning, our LoRD helps LDM generate high-quality images although LDM is fine-tuned on adversarial samples, demonstrating the effectiveness of our well-designed defense pipeline and LoRD modules. In addition, we realize the importance of fine-tuning clean samples, and present fine-tuning results for LDM using clean images in Fig.~\ref{fig:Clean samples finetuning for Face}. These results indicate LoRD pipeline can still performance well when fine-tuning clean samples.

\subsection{Quantitative Results}

We present our quantitative results in Table~\ref{fig:Main Quantitative Results}. These results demonstrate the efficacy of our defense method, which achieves higher CLIP-IQA scores and lower FID scores compared to ACE attacks in mist-v2 and adversarial fine-tuning using PGD-2. To further validate the effectiveness of our two-branch design and balance parameter $\lambda$, we conduct quantitative experiments on mixed data. As shown as Fig.~\ref{fig:mixed_results_merged}, we mix different proportions of adversarial samples with clean testing images datasets, and assess the CLIP-IQA and FID scores on mixed CelebA-HQ tesing images. The results demonstrate that our LoRD is robust to the change of proportion of adversarial samples, while standard LoRA fine-tuning is increasingly affected by the proportion of adversarial samples.

\section{Conclusion}
In this paper, we have presented a Low-Rank Defense (LoRD) method for defending against adversarial attacks on the LoRA fine-tuning for Latent Diffusion Models (LDMs). As an upgrade of the LoRA method, LoRD introduces the merging idea and balance parameter for fine-tuning an LDM on adversarial samples while maintaining its ability to handle clean samples. Extensive experiments on two facial image datasets and a landscape image dataset evaluate the effectiveness of our LoRD module and two-stage defense pipeline.

\bibliographystyle{IEEEbib}
\bibliography{icme2025references}

\end{document}